\documentclass{iosart2c}

\usepackage[T1]{fontenc}
\usepackage{times}%

\usepackage{t1enc}
\usepackage[latin1]{inputenc}

\usepackage{natbib}
\usepackage{amsmath}
\usepackage{dcolumn}
\usepackage{graphics,graphicx}

\usepackage{multirow}

\newcolumntype{d}[1]{D{.}{.}{#1}}

\firstpage{1} \lastpage{5} \volume{1} \pubyear{2010}

\begin{document}
\begin{frontmatter}                           % The preamble begins here.

%
%\pretitle{Pretitle}
\title{An Adaptive Fuzzy-Based System to Simulate, Quantify and Compensate Color Blindness}

\runningtitle{An Adaptive Fuzzy-Based System to Simulate, Quantify and Compensate Color Blindness}
%\subtitle{Subtitle}

\author[A]{\fnms{Jinmi} \snm{Lee}}
and
\author[B]{\fnms{Wellington} \snm{Pinheiro dos Santos}\thanks{Corresponding author. E-mail: wellington.santos@ieee.org}}
\runningauthor{J. Lee and W. P. dos Santos}
\address[A]{Escola Politécnica de Pernambuco, Universidade de Pernambuco,\\ Rua Benfica, 455, Madalena, Recife, Pernambuco, 50720-001, Brazil\\
E-mail: jinmilee@gmail.com}
\address[B]{Núcleo de Engenharia Biomédica, Universidade Federal de Pernambuco,\\ Av. Prof. Morais Rego, 1235, Cidade Universitária, Recife, Pernambuco, 50670-901, Brazil\\
E-mail: wellington.santos@ieee.org}

\begin{abstract}
About 8\% of the male population of the world are affected by a determined type of color vision disturbance, which varies from the partial to complete reduction of the ability to distinguish certain colors. A considerable amount of color blind people are able to live all life long without knowing they have color vision disabilities and abnormalities. Nowadays the evolution of information technology and computer science, specifically image processing techniques and computer graphics, can be fundamental to aid at the development of adaptive color blindness correction tools. This paper presents a software tool based on Fuzzy Logic to evaluate the type and the degree of color blindness a person suffer from. In order to model several degrees of color blindness, herein this work we modified the classical linear transform-based simulation method by the use of fuzzy parameters. We also proposed four new methods to correct color blindness based on a fuzzy approach: Methods A and B, with and without histogram equalization. All the methods are based on combinations of linear transforms and histogram operations. In order to evaluate the results we implemented a web-based survey to get the best results according to optimize to distinguish different elements in an image. Results obtained from 40 volunteers proved that the Method B with histogram equalization got the best results for about 47\% of volunteers.
\end{abstract}


\end{frontmatter}

\section{Introduction}

Millions of years of evolution have made the human visual system one of the most important sensory faculties. Nevertheless, about 8\% of the male population has some sort of color vision disturbance. This condition is characterized by the partial or complete reduction of the ability to distinguish some colors [1]. It is a relatively common fact that people with color blindness use to live several years without realizing that they have a color vision deficiency. The reason for this fact is that the disorder can appear in different intensities. The increasing user interaction with graphical interfaces has been evidencing several problems related to color discrimination, which often restrics the use of these web-based applications \cite{bruni2006}.

The increasing evolution of information technology and computer science, as well as digital image processing on the improvement of visual information for human interpretation, have improved the visual quality of digital images for people with certain degrees of disturbance in color perception. However, most applications intended to lessen the color blindness effects do not take in account that such disorders could occur in many degrees, depending on each person \cite{gonzalez2002}.

The classical methods to simulate color blindness are based on mathematical color models to represent extreme cases of color blindness, i.e. the absence of one of the three photoreceptors of the human eye (red, green, and blue, i.e. low, medium, and high frequencies) \cite{vienot1999}. Real cases of color blindness are characterized by some degree of anomaly at the absorbance spectrum of red, green, and blue spectral bands. Such a degree of chromatic deviation can be measured by software tools designed to collect parameters that can simulate real color blindness using fuzzy membership functions.

One of the targets of this paper is to perform a study on the chromatic abnormalities of the human visual system and to develop computational tools for adaptability of human-machine interfaces, providing the inclusion of individuals with color blindness and creating more accessible solutions.

In order to reach this target, we developed a simulation tool of color blindness from the application of linear transformation matrices that model the nonexistence or disability of the cone cells responsible for sensitivity to low, medium and high frequency spectral bands. We also developed a software tool to test color blindness and assess the degree of color blindness, using a fuzzy-based approach. The last tool was developed to improve the visual quality of digital images for people with color blindness. This integrated solution includes tests for the diagnosis of the color disturbance type to the image preview with the adjustments that aims to reduce the color blindness effects.

In order to contribute for the evaluation of future color blindness correction tools, another purpose of this paper is to present a fuzzy-based method to simulate real color blindness. The development of color blindness simulators is very important both for the development and design of GUIs (Graphical User Interfaces) and for supporting the development of new mathematical methods of correction of color blindness. We developed a tool which simulates color blindness from linear transformation matrices which model either the nonexistence or the disability of the cone cells responsible for the sensitivities to low, medium and high frequencies. Those matrices are adapted according to the fuzzy parameters obtained by a previous tool, briefly cited above, which evaluates color blindness and assesses its severity degree, using an approach based on Fuzzy Logic.

\section{Materials and Methods}

In the eyes, specifically in the retina, there is a zone where we can find the sensory cells specialized to capture the light stimuli: the photoreceptors known as rods and cones \cite{bruni2006}. The rods have a scotopic characteristic, i.e they have a high sensitivity to achromatic light. Since the cones have the photopic characteristic, they are less sensitive to light, but are able to discriminate between different wavelengths \cite{bruni2006}. There are three different types of cones in human eyes, each containing one type of photosensitive pigment. One type detects the spectrum of low frequencies of light (red color), while another one detects the medium frequencies spectrum (green color). The third type of cones detects the high frequencies spectral band (blue color). All these cones working together allow color vision \cite{bruni2006,vienot1999}.

The eye cones can be classified according to their sensitivity to different wavelengths. Those which are sensitive to the red spectral band are stimulated by long wavelengths; those ones sensitive to green are stimulated by wavelengths considered average, while cones that are sensitive to blue color are stimulated by short wavelengths. The vision of different types of colors becomes possible when those three types of cones are stimulated at the same time \cite{bruni2006}.

Several statistical studies show that about 8\% of males and 0.4\% of females have some form of disability related to the perception of colors \cite{bruni2006}. Deficiencies for the colors are also called dyschromatopsias, or simply color blindness \cite{vienot1999,bruni2006}.

Most people have trichromatic vision. However, regarding to color vision disorders such as color blindness, it is possible to classify them in \cite{bruni2006}:
\begin{description}
  \item [Anomalous trichromacy:] anomaly in the proportions of red, green and blue. There are three types of anomalous trichomatic classified:
  \begin{enumerate}
    \item Protanomaly: less sensitive to red.
    \item Deuteranomaly: less sensitive to green.
    \item Tritanomaly: less sensitive to colors in the range of blue-yellow.
  \end{enumerate}
  \item [Dichromatism:] it can be considered as special absolute cases of anomalous dichromatism, i.e. the total lack of sensitivity to red, green or blue. Due to the absence of one kind of cone, it is presented in the form of:
  \begin{enumerate}
    \item Protanopia: absence of red retinal photoreceptors.
    \item Deuteranopia: absence of green retinal photoreceptors.
    \item Tritanopia: absence of blue retinal photoreceptors.
  \end{enumerate}
  \item [Monochromatism:] Total inability to perceive color. This type of color blindness makes one see the world in gray levels. It is a very rare deficiency called achromatic vision.
\end{description}

Considering monochromatism as monospectral vision, it is not possible to realce image features depending on color. Furthermore, dichromatism can be perceived as extremes cases of anomalous trichromacy. Therefore, herein this paper we focused only on anomalous trichromacy and dichromatism.

In order to simplify notation and avoid transcribing the full name of the deficiencies of the chromatic visual system, throughout this article we decided to adopt the terms protan, deuteran and tritan to the deficiency of sensitivity to the frequencies of red (low frequencies), green (medium frequencies) and blue (high frequencies), respectively.

About twenty methods of diagnosis and classification of color disorders are most often used. Among them are: pseudoisochromatic plates or color discrimination tests, color arrangement tests or color hue test, equalization, appointment or designation, etc. Pseudoisochromatic plates are used in the discrimination tests. These various types of available plates are combined in various tests, among which the Ishihara test, that is quite popular, being the most known and used worldwide \cite{bruni2006}.

Although this test does not provide a quantitative assessment of the problem and does not identify deficiencies of the tritan type, studies show that the Ishihara test remains the most effective test for rapid identification of congenital deficiencies in color vision \cite{bruni2006}.

Once visual information depends on image color distribution, it also depends on the frequency response of the visual system. Consequently, the lack of information can constitute a considerable problem for people with color blindness. Thus, it is possible to say that color blindness constitutes an obstacle to the effective use of computers, which nowadays uses more and more graphics in its interface and its visual communication.

There is a growing commitment to create computational tools focused on the accessibility of people with color visual disturbances. The color blindness simulators are already quite common and avoid serious problems of accessibility, aiding to comprehend the perceptual limitations of a color blind individual \cite{vienot1999}. Nevertheless, these simulators are designed just for dichromatism \cite{vienot1999}.

There exist also applications designed to improve the visual quality of images. However, in most applications, it is assumed that the user already knows his type of disorder and does not consider that the disorder may occur in varying degrees. The difference of using an adaptive filter is related to the diagnosis and the use of an approach that considers the uncertainty associated with the problem, where Fuzzy Logic appears as a natural candidate to solve the adaptation to the user according to his degree of color blindness.

It is quite common for people with color blindness not to realize that they have visual disturbances, and many - when they discover the problem - do not know how to classify it. However, it is very important to improve efficiently the quality of life of those people, the knowledge of information as the type of color blindness and in what degree it is.

In order to fill this information gap, a test tool called DaltonTest was developed. The goal of this tool is to classify the color blindness, showing the degree of the disability and its possible forms of presentation.

\begin{figure}
  \centering
  \begin{minipage}[b]{0.48\linewidth}
    \centering
    \includegraphics[width=0.9\linewidth]{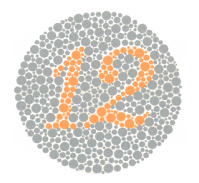}
    \\(a)\\
    \includegraphics[width=0.9\linewidth]{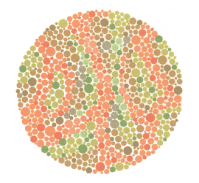}
    \\(b)\\
  \end{minipage}
  \begin{minipage}[b]{0.48\linewidth}
    \centering
    \includegraphics[width=0.9\linewidth]{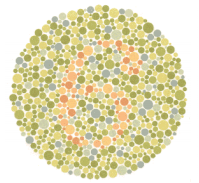}
    \\(c)\\
    \includegraphics[width=0.9\linewidth]{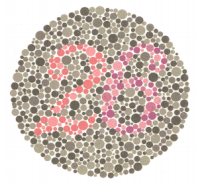}
    \\(d)\\
  \end{minipage}
  \caption{Examples of Ishihara plates: (a) demonstration plate; (b) hidden plate; (c) masked plate; and (d) diagnosis plate.}
  \label{fig:PranchasIshihara}
\end{figure}

In DaltonTest tool, the user is submitted to the Ishihara test (see figure \ref{fig:PranchasIshihara}), that has been customized through the use of weights. Different weights were assigned to the Ishihara test questions, so that the inaccurate responses received fewer points than the precise answers. This simple change makes it possible to evaluate approximately the user's blindness degree.
%\begin{figure}
%	\centering
%		\includegraphics[width=0.45\textwidth]{FigDaltonTest.png}
%	\caption{DaltonTest screen presenting a demonstration pseudoisochromatic plate}
%	\label{fig:FigDaltonTest}
%\end{figure}

\begin{figure}
	\centering
		\includegraphics[width=0.45\textwidth]{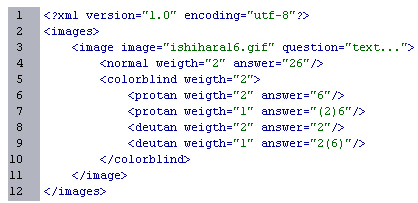}
	\caption{Example of XML code describing the data structure of the DaltonTest tool}
	\label{fig:FigDaltonTestXML}
\end{figure}

Figure \ref{fig:FigDaltonTestXML} shows an example of the data structure adopted to describe and store a question of the Ishihara test used in the DaltonTest tool. The test, when completed, presents an estimated diagnosis of the user color blindness, containing three factors: the degree of color blindness, the degree of protanomaly, and the degree of deuteranomaly. These degrees are in fact degrees of membership to fuzzy sets associated to types of color blindness, protanomaly, and deuteranomaly, calculated from the fuzzyfication of determined scores. Such result is then used by the correction tool, providing a fuzzy characteristic for the application.

The developed tool DaltonSim allows the simulation of the most common cases of anomalous trichomatic: protanomaly and the deuteranomaly.

The simulation algorithm of colorblindness is based on the color model LMS (Longwave, Middlewave, Shortwave), once this color model is the most adequate to model the behaviour of light reception: notice eye cones are organized in receptor groups of short, middle, and high frequencies \cite{vienot1999}. Conversion of the RGB components into components of the LMS model is the first step of the algorithm. The conversion is achieved by the application of a matrix, and, therefore, a linear conversion \cite{bruni2006}:
\begin{equation}
  \left[
  \begin{array}{l}
  L\\
  M\\
  S\\
  \end{array}
  \right]
  =
  T_{RGB-LMS}
  \left[
  \begin{array}{l}
  R\\
  G\\
  B\\
  \end{array}
  \right],
\end{equation}

\begin{equation}
  \left[
  \begin{array}{l}
  L\\
  M\\
  S\\
  \end{array}
  \right]
  =
  \left[
  \begin{array}{ccc}
    {17.8824} & {43.5161} & {4.1194}\\
    {3.4557} & {27.1554} & {3.8671}\\
    {0.0300} & {0.1843} & {1.4671}\\
  \end{array}
  \right]
  \left[
  \begin{array}{l}
  R\\
  G\\
  B\\
  \end{array}
  \right].
\end{equation}

The second step is reducing the normal domain of colors to the domain of a color blind individual. The linear transformation for protanopia is expressed as follows \cite{vienot1999}:
\begin{equation} \label{eq:absprotan}
  \left[
  \begin{array}{l}
  L_p\\
  M_p\\
  S_p\\
  \end{array}
  \right]
  =
  \left[
  \begin{array}{ccc}
    {0} & {2.0234} & {-2.5258}\\
    {0} & {1} & {0}\\
    {0} & {0} & {1}\\
  \end{array}
  \right]
  \left[
  \begin{array}{l}
  L\\
  M\\
  S\\
  \end{array}
  \right],
\end{equation}
and for deuteranopia \cite{vienot1999}:
\begin{equation} \label{eq:absdeuteran}
  \left[
  \begin{array}{l}
  L_d\\
  M_d\\
  S_d\\
  \end{array}
  \right]
  =
  \left[
  \begin{array}{ccc}
    {1} & {0} & {0}\\
    {0.4942} & {0} & {1.2483}\\
    {0} & {0} & {1}\\
  \end{array}
  \right]
  \left[
  \begin{array}{l}
  L\\
  M\\
  S\\
  \end{array}
  \right].
\end{equation}

Finally, there must be a transformation of the LMS color model to RGB. This transformation is obtained using the inverse matrix of the first step matrix \cite{vienot1999}.
\begin{equation}
  \left[
  \begin{array}{l}
  R\\
  G\\
  B\\
  \end{array}
  \right]
  =
  T_{LMS-RGB}
  \left[
  \begin{array}{l}
  L\\
  M\\
  S\\
  \end{array}
  \right],
\end{equation}

\begin{equation}
  \left[
  \begin{array}{l}
  R\\
  G\\
  B\\
  \end{array}
  \right]
  =
  T_{RGB-LMS}^{-1}
  \left[
  \begin{array}{l}
  L\\
  M\\
  S\\
  \end{array}
  \right],
\end{equation}

\begin{equation}
  \left[
  \begin{array}{l}
  R\\
  G\\
  B\\
  \end{array}
  \right]
  =
  \left[
  \begin{array}{ccc}
    {0.0809} & {-0.1305} & {0.1167}\\
    {-0.0102} & {0.0540} & {-0.1136}\\
    {-0.0004} & {-0.0041} & {0.6935}\\
  \end{array}
  \right]
  \left[
  \begin{array}{l}
  L\\
  M\\
  S\\
  \end{array}
  \right].
\end{equation}

In order to modify these dichromatism models, we generate other matrices by including several fuzzy parameters, in order to get a mathematical model useful to deal with anomalous trichromatism. Therefore, our fuzzy-based model considers not only the extreme cases of protanopia and deuteranopia, but also hybrid cases where each case of color reception, including color ``normality'', is represented by a given degree of membership to fuzzy sets designed to model these cases \cite{juang2004, zhu2003, axer2003, wang2004, hung2006, alexiuk2005, yager2000, lucas1999, villar2009, rokni2010, smith2007, lee2007, sabourin2007, perusich2008, adeli1994, sarma2000a, sarma2000b, adeli2000, samant2001, karim2002, sarma2002, jiang2003, adeli2003, adeli2006, jiang2008, reuter2010, kim2010, sadeghi2010, bianchini2010, bianchini2010, jin2009}.

The fuzzy parameter were designed to generate a linear dichromatism model to get LMS models able to vary from nondichromatic matrices represented by identity matrices to the absolute dichromatic simulation matrices given by expressions \ref{eq:absprotan} and \ref{eq:absdeuteran}. Consequently, for the fuzzy degree of protanomaly $\alpha_p$, if $\alpha_p=0$ we have the identity matrix, whilst in case we have $\alpha_p=1$, we get the transform matrix described by expression \ref{eq:absprotan}. The analysis is similar for the deuteranopia simulation matrix. Our proposal deals with the generation of intermediate simulation matrices able to model nonabsolute cases of dichromatism. Therefore, considering the components already converted into the LMS model, the linear transformation for pro\-ta\-no\-ma\-ly is proposed as following:
\begin{equation}
  \left[
  \begin{array}{l}
  L_p\\
  M_p\\
  S_p\\
  \end{array}
  \right]
  =
  \left[
  \begin{array}{ccc}
    {(1-\alpha_p)} & {2.0234\alpha_p} & {-2.5258\alpha_p}\\
    {0} & {1} & {0}\\
    {0} & {0} & {1}\\
  \end{array}
  \right]
  \left[
  \begin{array}{l}
  L\\
  M\\
  S\\
  \end{array}
  \right],
\end{equation}
where $\alpha_p$ is equal to the degree of protan; and for deuteranomaly:
\begin{equation}
  \left[
  \begin{array}{l}
  L_d\\
  M_d\\
  S_d\\
  \end{array}
  \right]
  =
  \left[
  \begin{array}{ccc}
    {1} & {0} & {0}\\
    {0.4942\alpha_d} & {(1-\alpha_d)} & {1.2483\alpha_d}\\
    {0} & {0} & {1}\\
  \end{array}
  \right]
  \left[
  \begin{array}{l}
  L\\
  M\\
  S\\
  \end{array}
  \right],
\end{equation}
where $\alpha_d$ is equal to the degree of deuteran.

Based on both linear transformations, we propose a hybrid model of protanomaly and deuteranomaly:
\begin{equation}
  \left[
  \begin{array}{l}
  L_h\\
  M_h\\
  S_h\\
  \end{array}
  \right]
  =
  \left[
  \begin{array}{ccc}
    {(1-\alpha_p)} & {2.0234\alpha_p} & {-2.5258\alpha_p}\\
    {0.4942\alpha_d} & {(1-\alpha_d)} & {1.2483\alpha_d}\\
    {0} & {0} & {1}\\
  \end{array}
  \right]
  \left[
  \begin{array}{l}
  L\\
  M\\
  S\\
  \end{array}
  \right].
\end{equation}

The degrees of protanomaly $\alpha_p$ and deuteranomaly $\alpha_d$ are parameters obtained from the testing tool DaltonTest. Therefore, it is possible to simulate real cases of color blindness.

The DaltonSim has a very simple interface, and the result from the simulation is the visualization of the original images alongside the simulated images.
\begin{figure}
  \centering
  \begin{minipage}[b]{0.32\linewidth}
    \centering
    \includegraphics[width=0.95\linewidth]{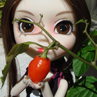}
    \\(a)\\
  \end{minipage}
  \begin{minipage}[b]{0.32\linewidth}
    \centering
    \includegraphics[width=0.95\linewidth]{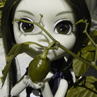}
    \\(a)\\
  \end{minipage}
  \begin{minipage}[b]{0.32\linewidth}
    \centering
    \includegraphics[width=0.95\linewidth]{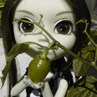}
    \\(a)\\
  \end{minipage}
  \caption{Result of dichromate simulation. (a) Original image, normal vision. (b) Image simulating protan type color blindness. (c) Image simulating type deuteran color blindness.}
  \label{fig:DichromateSimularion}
\end{figure}

The simulation has shown to be essential to understand the problems of accessibility of the color blind individuals. Figure \ref{fig:DichromateSimularion} shows some results from the simulation tool. Notice the presence of green color, since the absence of cones that detect green, in this case the type deuteran color blindness, does not prevent the perception of this spectral range, as there is a compensation due to the presence of other cones, as evidenced in the mathematical model of the anomaly deuteran. However, there is no perception of the real green \cite{vienot1999}.

Herein this work we also implemented a software tool called DaltonCor, intended to improve the visual quality of the color blind individuals, whereas its deficiency may be presented in different degrees. Two methods (A and B) were developed for the DaltonCor tool. Method A is based on the generation of a compensated image generated by the linear combination of the original image and the corrected versions for protan and deuteran cases. The weights are based on fuzzy rules. Method B is based on a linear transform matrix based on the fuzzy degrees of protan and deuteran color blindness, without using additional fuzzy rules. Both methods try to compensate for the lack of sensitivity to a particular color with new values for the normal color perception.

\subsection{Method A}

The Method A was divided into three modules: Filter, Fuzzy and Control. This last one, in addition to link the other two modules, communicates with the graphical interface and feeds it with information. The solution developed in the filter module is based on color perception in cases of absolute color blindness. First, let us focus on the protan color blindness:

Consider $f=(f_r,f_g,f_b)$ the original image followed by its three bands of color $r$, $g$ and $b$. Its correction is performed by following two steps. The first one (equation \ref{eq:PassoAp1}) is to assign new values to the bands not affected by color blindness. To protans, these bands of color are $f_g$ and $f_b$, as following:
\begin{equation} \label{eq:PassoAp1}
  f'=(f_r,f_g',f_b'),
  \left\{
  \begin{array}{l}
    {f_g'=\frac{1}{2}(f_r+f_g)}\\
    {f_b'=\frac{1}{2}(f_r+f_b)}\\
  \end{array}
  \right..
\end{equation}

In order to increase the visual quality of the image, the second step (equation \ref{eq:PassoAp2}) is intended to improve the contrast. The chosen technique was the standard histogram equalization presented by Gonzalez and Woods (2002) \cite{gonzalez2002}. Considering $\gamma$ the contrast optimizer (histogram equalizer), we have the following expression:
\begin{equation} \label{eq:PassoAp2}
  f_p=(f_r,f_g'',f_b''),
  \left\{
  \begin{array}{l}
    {f_g''=\gamma(f_g')}\\
    {f_b''=\gamma(f_b')}\\
  \end{array}
  \right..
\end{equation}

The correction equations for deuterans (\ref{eq:PassoAd1} and \ref{eq:PassoAd2}) are similar to those presented above:
\begin{equation} \label{eq:PassoAd1}
  f'=(f_r',f_g,f_b'),
  \left\{
  \begin{array}{l}
    {f_r'=\frac{1}{2}(f_r+f_g)}\\
    {f_b'=\frac{1}{2}(f_g+f_b)}\\
  \end{array}
  \right..
\end{equation}

\begin{equation} \label{eq:PassoAd2}
  f_d=(f_r'',f_g,f_b''),
  \left\{
  \begin{array}{l}
    {f_r''=\gamma(f_r')}\\
    {f_b''=\gamma(f_b')}\\
  \end{array}
  \right..
\end{equation}

The Filter Module returns two corrected images, $f_p$ and $f_d$. The Fuzzy Module is responsible for the filter customization and, based on the outcome from the test tool, assigns a fuzzy character of this correction. From an experimental approach, the following fuzzy rules were proposed, based on the output data we collect from DaltonTest color blindness software test \cite{lucas1999,yager2000,fan2002,lee2010a}:
\begin{equation}
  x_p'=\beta\wedge\alpha_p,
\end{equation}
where $x_p'$ is equal to the degree of color blindness $\beta$ with conjunction the degree of protan $\alpha_p$.

\begin{equation}
  x_d'=\beta\wedge\alpha_d,
\end{equation}
where $x_d'$ is equal to the degree of color blindness $\beta$ with conjunction the degree of deuteran $\alpha_d$.

\begin{equation}
  x_n'=\alpha_n \wedge (\neg\beta),
\end{equation}
where $x_n'$ is equal to the degree of normality $\alpha_n$ and in conjunction with not color blindness.

From the measurements obtained from the rules above, it is possible to make a fuzzification process on these magnitudes to obtain the weights expressed by the following equations:
\begin{equation}
  x_p=\frac{x_p'}{x_p'+x_d'+x_n'},
\end{equation}
\begin{equation}
  x_d=\frac{x_p'}{x_p'+x_d'+x_n'},
\end{equation}
\begin{equation}
  x_n=\frac{x_p'}{x_p'+x_d'+x_n'}.
\end{equation}

The corrected image $f^{*}$ is a weighted average of the corrected images for protan and deuteran type color blindness, and the original image, as represented in the following expression:
\begin{equation}
  f^{*}=x_p f_p+x_d f_d+ x_n f.
\end{equation}

\subsection{Method B}

The Method B is based on linear transformations designed to adaptatively correct the effects of chromatic disturbance. Differently from method A, this method is divided into two modules: Correction and Control. It is important to notice that, in this method, there is no previous correction for absolute color blindness, as it occurs in the module Filter of Method A. The module Control furnishes the graphic interface important information to be described in the following paragraphs.

The solution implemented in module Correction is inspired by the perception of colors in absolute color blindness. However, instead of using the absolute degrees of color blindness, we take into account the fuzzy degrees of protanomaly and deuteranomaly. Therefore, this method is an adaptative correction method.

Similarly to the method A, the proposal of correction of method B also tries to compensate the lack of sensitivity for determined colors (in fact, the deficiency to receive determined spectral bands), increasing their occurrence in the digital image and modifying the colors acquired with normal perception.

Let $f=(f_r,f_g,f_b)$ be the original image with its three color bands: $r$, $g$ and $b$. This image is compensated by the use of a transform matrix to correct the pixel values by changing them using linear combinations. The expressions \ref{eq:PassoBp} and \ref{eq:PassoBd} represent the corrections for the absolute cases of protan and deuteran using fuzzy features. These empirical expressions are based on the following idea: compensating the lack of color band in color blind vision by equally distributing the information of this band to the other two bands. For protans, the corrected image is proposed as following:
\begin{equation} \label{eq:PassoBp}
  f_p=(f_r,f_g',f_b'),
  \left\{
  \begin{array}{l}
    {f_g'=\frac{\alpha_p}{2}f_r+\frac{2-\alpha_p}{2}f_g}\\
    {f_b'=\frac{\alpha_p}{2}f_r+\frac{2-\alpha_p}{2}f_b}\\
  \end{array}
  \right.;
\end{equation}
for deuterans, the customization is performed in a similar manner, as can be seen in expression \ref{eq:PassoBd}:
\begin{equation} \label{eq:PassoBd}
  f_d=(f_r',f_g,f_b'),
  \left\{
  \begin{array}{l}
    {f_r'=\frac{\alpha_d}{2}f_g+\frac{2-\alpha_d}{2}f_r}\\
    {f_b'=\frac{\alpha_d}{2}f_g+\frac{2-\alpha_d}{2}f_b}\\
  \end{array}
  \right..
\end{equation}

In order to deal with hybrid cases of color blindness where the individual has anomalous trichromatism for the red and green spectral bands at the same time, the new values of bands $f_r$, $f_g$ and $f_b$ are calculated as following:
\begin{equation}
  f^{*}=(f_r',f_g',f_b'),
\end{equation}
\begin{equation}
  f_r'=\frac{\alpha_d}{2}f_g+\frac{2-\alpha_d}{2}f_r
\end{equation}
\begin{equation}
  f_g'=\frac{\alpha_p}{2}f_r+\frac{2-\alpha_p}{2}f_g
\end{equation}
\begin{equation}
  f_b'=\frac{\alpha_p}{4}f_r+\frac{\alpha_d}{4}f_g+\frac{4-\alpha_p-\alpha_d}{4}f_b
\end{equation}

Based on the new values of the three spectral bands, $r$, $g$ and $b$, it was possible to generate an unique transform matrix to deal with the adaptative correction of anomalous trichromatism cases, as seen in expression \ref{eq:MTransformB}:
\begin{equation} \label{eq:MTransformB}
  \left[
  \begin{array}{l}
  R'\\
  G'\\
  B'\\
  \end{array}
  \right]
  =
  \left[
  \begin{array}{ccc}
    {1-\frac{\alpha_d}{2}} & {\frac{\alpha_d}{2}} & {0}\\
    {\frac{\alpha_p}{2}} & {1-\frac{\alpha_p}{2}} & {0}\\
    {\frac{\alpha_p}{4}} & {\frac{\alpha_d}{4}} & {1-\frac{\alpha_p+\alpha_d}{4}}\\
  \end{array}
  \right]
  \left[
  \begin{array}{l}
  R\\
  G\\
  B\\
  \end{array}
  \right].
\end{equation}

The result of the application of the transform matrix is an image modified based on the given degrees of protan and deuteran.

\section{Results}

\subsection{Simulation Results}

The fuzzy-based simulation tool presented in this paper receives as inputs the results obtained by the customized Ishihara testing tool, DaltonTest. These parameters allow our simulation tool to generate results in accordance to what is expected from real color blindness cases, which is essential to understand how people with different degrees of color blindness perceive color variation.
 
\begin{figure}
  \centering
  \begin{minipage}[b]{0.32\linewidth}
    \centering
    PROTAN\\
    \includegraphics[width=0.9\linewidth]{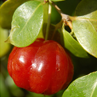}
    \\(b)\\
    \includegraphics[width=0.9\linewidth]{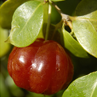}
    \\(c)\\
    \includegraphics[width=0.9\linewidth]{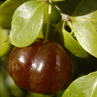}
    \\(d)\\
    \includegraphics[width=0.9\linewidth]{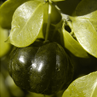}
    \\(e)\\
  \end{minipage}
  \begin{minipage}[b]{0.32\linewidth}
    \centering
    ORIGINAL\\
    \includegraphics[width=0.9\linewidth]{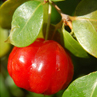}
    \\(a)\\
    DEUTERAN\\
    \includegraphics[width=0.9\linewidth]{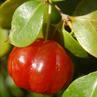}
    \\(f)\\
    \includegraphics[width=0.9\linewidth]{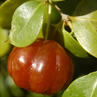}
    \\(g)\\
    \includegraphics[width=0.9\linewidth]{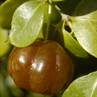}
    \\(h)\\
    \includegraphics[width=0.9\linewidth]{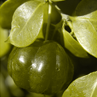}
    \\(i)\\
  \end{minipage}
  \begin{minipage}[b]{0.32\linewidth}
    \centering
    HYBRID\\
    \includegraphics[width=0.9\linewidth]{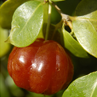}
    \\(j)\\
    \includegraphics[width=0.9\linewidth]{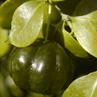}
    \\(k)\\
    \includegraphics[width=0.9\linewidth]{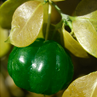}
    \\(l)\\
    \includegraphics[width=0.9\linewidth]{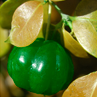}
    \\(m)\\
  \end{minipage}
  \caption{Results of dichromate simulation: (a) original image, normal vision; (b), (c), (d) and (e): protan type color blindness, for degrees of 25\%, 50\%, 75\%, and 100\%, respectively; (f), (g), (h) and (i): deuteran type color blindness, for degrees of 25\%, 50\%, 75\%, and 100\%, respectively; (j), (k), (l) and (m): hybrid (protan and deuteran) type color blindness, for equal degrees of protan and deuteran of 25\%, 50\%, 75\%, and 100\%, respectively.}
  \label{fig:SimulationResults}
\end{figure}

Notice in figure \ref{fig:SimulationResults} a simulation for different degrees of the color vision anomalies and the gradual loss of ability to distinguish different colors. 

It is important to notice the similarity of images figure \ref{fig:SimulationResults}(c) is a protanomaly in 50\% of cones and figure \ref{fig:SimulationResults}(j) is a hybrid case: 25\% protanomaly and 25\% deuteranomaly, which together reduce the compensation due to the presence of other cones, as can be perceived in the new mathematical model we presented herein this work.

The simulation results with different levels of abnormality proved to be very satisfactory, with exception of hybrid cases with more than 50\% of protan and deuteran. %The proposed model, however, still needs to be adjusted to avoid the excessive compensation of color.

\begin{figure}
  \centering
  \begin{minipage}[b]{0.48\linewidth}
    \centering
    ORIGINALS\\
    \includegraphics[width=0.9\linewidth]{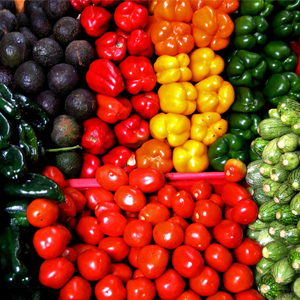}
    \\(a)\\
    \includegraphics[width=0.9\linewidth]{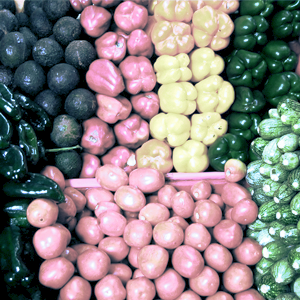}
    \\(c)\\
    \includegraphics[width=0.9\linewidth]{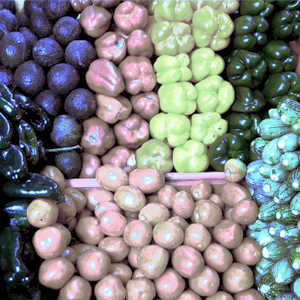}
    \\(e)\\
  \end{minipage}
  \begin{minipage}[b]{0.48\linewidth}
    \centering
    SIMULATED\\
    \includegraphics[width=0.9\linewidth]{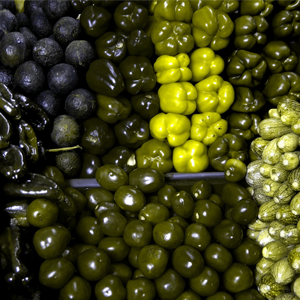}
    \\(b)\\
    \includegraphics[width=0.9\linewidth]{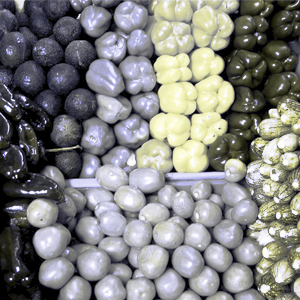}
    \\(d)\\
    \includegraphics[width=0.9\linewidth]{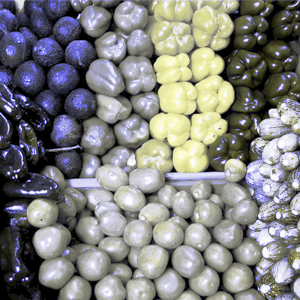}
    \\(f)\\
  \end{minipage}
  \caption{Results from image correction using the methods A and B in RGB with histogram equalization: (a) original image; (b) simulation for the absolute protan type; (c) corrected image with Method A; (d) corrected image with Method A simulated for protan color blindness; (e) corrected image with Method B; (f) corrected image with Method B simulated for protan color blindness.}
  \label{fig:CorrectionResults}
\end{figure}

In order to analyze the results of the correction tool proposed, 10 images in bitmap 32 bits format were used. The use of the simulation tool was essential to understand how the corrections would be perceived by people with color blindness. We also analyzed all the correction variations in RGB, LMS, and with and without histogram equalization. In figure \ref{fig:CorrectionResults}, it can be seen the correction result for an individual 100\% color blind, 0\% protan, 100\% deuteran and 0\% normal.

Notice that in figure \ref{fig:CorrectionResults}(b) the red tomatoes and the green peppers appear nearly equal. However, in figure \ref{fig:CorrectionResults}(d) the shades of tomatoes and peppers are very different. Thus it is easy to see the gain of visual information after correcting the image.

\subsection{Results for Color Blind People}

Although the analysis done by the simulator display is satisfactory, a group of four people with color blindness volunteered to test the tool. The volunteers were evaluated with the customized Ishihara test using the test tool. The test set was composed by a set of 32 bits bitmap images with size of 300$\times$300 distributed as following: 10 original images without correction and, for each of these images, 4 corrected images using Method A with and without histogram equalization, taking into account RGB and LMS images, composing a total of 44 images. Through the obtained results, the four combinations of correction were analyzed. The results are presented in Table \ref{tab:ResultsCorrectionFilters}. Althoug these results were not conclusive due to the small number of volunteers, it was important as a first feedback from color blind people.

Each corrected image was subjectively evaluated and judged as \emph{much better}, \emph{better}, \emph{indifferent}, \emph{worse} or \emph{much worse}. Distortion and greater ability to distinguish elements in relation to the original image were considered as evaluation criteria. 

It was observed that, by the correction in RGB domain with histogram equalization, the images became more understandable, because elements that were perceived with the same color (due to color blindness) received different colors. Some images after the correction showed less saturation in their colors and some of these cases were considered worse. With the correction in RGB without histogram equalization there was not a big gain in the visual improvement of images, although this type of correction has been found to cause less negative impact on the distortion of the original colors.

The correction in LMS, in general, changed overmuch the original colors of the images, although it performed efficiently on images that purposely have hidden elements, like the images used in pseudoisochromatic Ishihara plates.
\begin{table}
	\centering
		\begin{tabular}{c|l|l|l|l}
			\hline
			{} & \multicolumn{2}{c|}{With equalization} & \multicolumn{2}{c}{Without equalization}\\
			\hline
			\multirow{5}{*}{RGB} & {Much better} & {20\%} & {Much better} & {0\%}\\
			{} & {Better} & {46\%} & {Better} & {43\%}\\
			{} & {Indifferent} & {20\%} & {Indifferent} & {43\%}\\
			{} & {Worse} & {14\%} & {Worse} & {14\%}\\
			{} & {Much worse} & {0\%} & {Much worse} & {0\%}\\
			\hline
			\multirow{5}{*}{LMS} & {Much better} & {20\%} & {Much better} & {7\%}\\
			{} & {Better} & {17\%} & {Better} & {33\%}\\
			{} & {Indifferent} & {20\%} & {Indifferent} & {3\%}\\
			{} & {Worse} & {7\%} & {Worse} & {7\%}\\
			{} & {Much worse} & {36\%} & {Much worse} & {50\%}\\
			\hline			
		\end{tabular}
	\caption{Preliminar evaluation by a set of 4 volunteers of 40 images corrected using Method A based on RGB and LMS, with and without histogram equalization}
	\label{tab:ResultsCorrectionFilters}
\end{table}

\subsection{Results for People without Color Blindness}

Considering that: 1) the amount of people with color blindness represents a considerably lower percentage in the world's population \cite{vienot1999,bruni2006}, 2) the occurrence of color blindness is not related to factors that might allow a clear separation between color blind individuals and the rest of the population, it is quite difficult to assemble a good set of test tools to validate the correction of color blindness proposed in this paper. Thus, we build a test based on the web to collect information of a subjective nature that could be useful to distinguish the methods proposed in this paper according to a qualitative evaluation. 

In order to build the platform test site we used 10 basic images. For each of these 10 images we generated images for protan and deuteran color blindness using the simulation method proposed in this work, with degrees of color blindness of 0\%, 25\%, 50\%, 75\% and 100\%. We also used images with 0\% of blindness as a way of maintaining quality control of results. Thus, we generated a total amount of 450 images.

The web survey consists of 90 questions with five randomly arranged choices. Each issue consists in presenting an image, converted to the vision of a colorblind using four simulation methods proposed, with a certain degree of colorblindness. For each presented image there are five options, where four of them are obtained from the application of the correction methods, while the other image is simply a copy of the presented image. The user must choose the option among five options that best highlights different elements in the presented image, i.e. the option that optimizes the contrast between different elements, even if these elements are not clearly distinguished in the original image.

\begin{figure}
	\centering
		\includegraphics[width=0.45\textwidth]{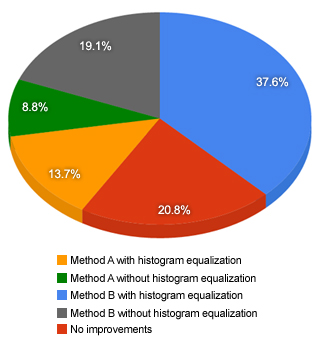}
	\caption{Percentages of responses versus color blindness correction methods, considering a set of 40 volunteers without color blindness}
	\label{fig:G1_Agrupar_Todos}
\end{figure}

\begin{figure}
	\centering
		\includegraphics[width=0.45\textwidth]{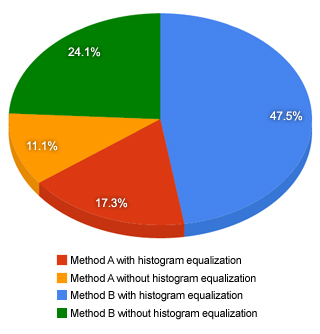}
	\caption{Percentages of positive responses (responses different from ``no improvements'') versus color blind correction methods, considering a set of 40 volunteers without color blindness}
	\label{fig:G2_Somente_Metodos}
\end{figure}

The web survey was administered to a universe of 40 users without color blindness. The absolute results are present in figure \ref{fig:G1_Agrupar_Todos}. From the results presented, we can notice that subjective aesthetic criteria influenced the results: 20.8\% of volunteers chose the option ``No improvements'', when the expected number was 11.1\% (10 of the 90 presented images with 0\% of color blindness). Figure \ref{fig:G2_Somente_Metodos} shows the results without the option ``No improvements''.

\begin{figure}
	\centering
		\includegraphics[width=0.45\textwidth]{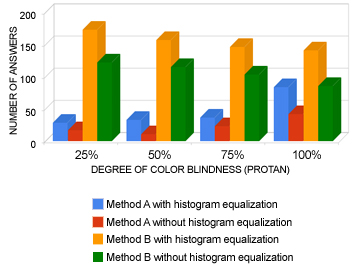}
	\caption{Percentages of positive responses (responses different from ``no improvements'') versus the degree of protan color blindness, considering a set of 40 volunteers without color blindness}
	\label{fig:G3_Protan}
\end{figure}

\begin{figure}
	\centering
		\includegraphics[width=0.45\textwidth]{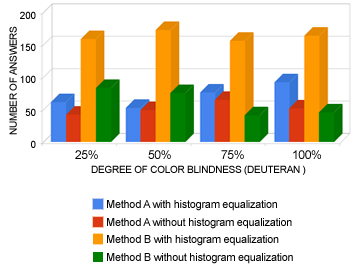}
	\caption{Percentages of positive responses (responses different from ``no improvements'') versus the degree of deuteran color blindness, considering a set of 40 volunteers without color blindness}
	\label{fig:G3_Deuteran}
\end{figure}

The results for protan and deuteran colorblindness distributed according to the degree of color blindness are shown in figures \ref{fig:G3_Protan} and \ref{fig:G3_Deuteran}, respectively.

Figure \ref{fig:G3_Protan} shows that Method B with histogram equalization received the greatest number of positive responses from volunteers. Method B without histogram equalization also had a reasonable amount of positive responses for all degrees of color blindness, yet its performance was inferior to that obtained by Method B with histogram equalization. Method A, with and without histogram equalization, is shown below the Method B, for the degrees of color blindness of 25\%, 50\% and 75\%. However, for protans with 100\% of color blindness, Method A with histogram equalization can achieve a similar performance to the performance of Method B without histogram equalization. For protans with 100\% color blindness, Method A without histogram equalization reaches almost twice the performance obtained for 25\% and 50\% protans. However, despite the improvement acquired for protan cases with 100\% of color blindness, Method A without histogram equalization is still much lower than other methods.

\begin{figure}
	\centering
		\includegraphics[width=0.45\textwidth]{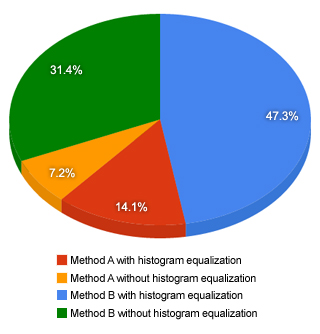}
	\caption{Percentages of positive responses (responses different from ``no improvements'') versus color blind correction methods, considering a set of 40 volunteers without color blindness and protan simulated images}
	\label{fig:G2_1_Anomalia_Protan}
\end{figure}

Figure \ref{fig:G2_1_Anomalia_Protan} shows the overall results for protan colorblindness. The results show that, for protan colorblindness, Method B with histogram equalization had the best performance: 47.3\% of positive responses, compared with Method B without histogram equalization, with 31.4\% of positive responses. Method A was very inferior to Method B, with 14.1\% and 7.2\% of positive responses, with and without histogram equalization, respectively. The results also show that the histogram equalization is also an important factor in improved distinction of different elements in the images.

Figure \ref{fig:G3_Deuteran} shows the results for deuteran color blindness, considering all degrees of color blindness. Here again the results obtained with Method B with histogram equalization were superior to those obtained by other methods. However, the distance to Method B without histogram equalization increased. The performance of Method A, with and without histogram equalization, improved slightly to 25\%, 50\% and 75\% of blindness in relation to protan cases, although it remains well below the Method B with histogram equalization, but the behavior for 100\% of color blindness was similar. Unlike the protan cases, Method B without histogram equalization had the worst performance among the four methods for 75\% of blindness.

\begin{figure}
	\centering
		\includegraphics[width=0.45\textwidth]{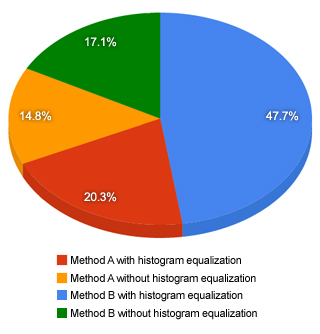}
	\caption{Percentages of positive responses (responses different from ``no improvements'') versus color blind correction methods, considering a set of 40 volunteers without color blindness and deuteran simulated images}
	\label{fig:G2_2_Anomalia_Deuteran}
\end{figure}

Figure \ref{fig:G2_2_Anomalia_Deuteran} illustrates the general results for the cases of deuteran color blindness. Comparing the results for deuterans with the results for protans, it is clear to notice that the performance of Method B with histogram equalization remained virtually the same, at approximately 47\%. However, Method A with and without histogram equalization had much better results for deuterans than for protans. Already Method B without histogram equalization proved to be far worse for deuterans than for protans. Therefore it is important to notice that there is a clear improvement of results due to the addition of histogram equalization.

\section{Discussion and Conclusion}

Despite the rapid technological developments and advancements in the area of digital image processing, it is not easy to find tools that simulate milder, intermediate forms of color vision impairment and reduce the effects of chromatic visual disturbances. As cited before, the extreme forms of color blindness - which are characterized by the complete functional absence of one type of cones - are not as common as the milder forms with partial or shifted sensitivity. However, many studies assume that if a color scheme is legible for someone with extreme color vision impairment, it will also be easily legible for those with a minor affliction, and ignore the possibility of adaptive and customized correction methods.

This work presents an adaptive tool to simulate color blindness. The fuzzy parameters designed to make the simulation flexible are obtained from the testing tool previously presented. 

Therefore, this paper proposed the development of a set of computational tools to improve the accessibility and visual quality of life for color blind people. However, one of the greatest difficulties encountered in developing this work is the relative lack in the literature about other mathematical methods for adaptive compensation of color blindness as the proposed.

As mentioned, color blindness affects only a low percentage of population. The absence of color blindness cases in a statistically significant amount can be an obstacle to the development of adaptive tools for correction of the anomaly, especially when the focus is on real cases, rather than in extreme cases (dichromatism). Thus good color blindness simulators are of great importance on generating of synthetic cases of color vision disturbance in a statistically significant amount, so that they can be used in the development of correction tools. The fuzzy-based simulation proved to be essential to understand the accessibility problems of people with red and green color disorders in different degrees.

Tests were conducted with a group of people with color blindness. This experience was important to evaluate the results obtained from the correction filters, and to more accurately gauge the difficulties encountered by the volunteers. The results were very positive, confirming that the proposed correction is able to extract a higher amount of information from an image.

Moreover, all the tools proved to be intuitive and easy to use, providing a better user experience, and consequently their habitual use.

Extensive tests were also conducted with people without color blindness using color blindness simulators and a web tool based on a survey. These tests were performed by a group of 40 volunteers. Results indicated that the Method B with histogram equalization got better results for both protan and deuteran color blindness.

% Referências bibliográficas
\bibliographystyle{plain}
\bibliography{arq_bib}

\end{document}